%% file: main.tex
\documentclass{article}
\usepackage{spconf,amsmath,graphicx}
\usepackage{algorithm}
\usepackage{algorithmic}
\usepackage{booktabs}
\usepackage{amssymb}

\title{GenGMM: Generalized Gaussian-Mixture-based Domain Adaptation Model for Semantic Segmentation}
\name{Nazanin Moradinasab$^{1 \star}$\thanks{$^{\star}$Email: nm4wu@virginia.edu\\
This study is supported by the National Center for Advancing Translational Science of the National Institutes of Health Award UL1TR003015/ KL2TR003016.} \qquad Hassan Jafarzadeh $^{1}$ \qquad Donald E. Brown$^{2}$}
\address{$^{1}$ Department of Engineering Systems and Environment, University of Virginia, Charlottesville, VA, USA\\ 
\qquad{$^{2}$ School of Data Science, University of Virginia, Charlottesville, VA, USA}} 

\begin{document}
%
\maketitle
\begin{abstract}
Domain adaptive semantic segmentation is the task of generating precise and dense predictions for an unlabeled target domain using a model trained on a labeled source domain. While significant efforts have been devoted to improving unsupervised domain adaptation for this task, it is crucial to note that many models rely on a strong assumption that the source data is entirely and accurately labeled, while the target data is unlabeled. In real-world scenarios, however, we often encounter partially or noisy labeled data in source and target domains, referred to as Generalized Domain Adaptation (GDA). In such cases, we suggest leveraging weak or unlabeled data from both domains to narrow the gap between them, resulting in effective adaptation. We introduce the Generalized Gaussian-mixture-based (GenGMM) domain adaptation model, which harnesses the underlying data distribution in both domains to refine noisy weak and pseudo labels. The experiments demonstrate the effectiveness of our approach.
\end{abstract}
\section{INTRODUCTION}
\label{sec:intro}
\input{introduction}

\section{METHODOLOGY}
\input{methodology}

\section{EXPERIMENTS}
\label{sec:exp}
\input{experiments}


\section{CONCLUSION}
\input{conclusion}

\section{Acknowledgement}
This work was supported by the National Center for Advancing Translational Science of the National Institutes of Health Award UL1 TR003015,  R01DK131491, and an American Heart Association
Predoctoral Fellowship (23PRE1028980).



\bibliographystyle{IEEEbib}
\bibliography{strings,refs}

\end{document}

%% file: introduction.tex
The success of deep learning models relies on acquiring abundant large annotated datasets \cite{cordts2016cityscapes}. However, annotating such datasets for semantic segmentation is costly and time-intensive. To alleviate this burden, Unsupervised Domain Adaptation (UDA) techniques \cite{hoyer2022daformer, xie2023sepico} seek to harness knowledge from a labeled source domain to improve learning in an unlabeled target domain.
However, UDA methods assume fully labeled source data and completely unlabeled target domains, which is often not the case in practice (e.g. healthcare, or autounoumous) due to partial or noisy labels in both domains. 
Therefore, we introduce a novel domain adaptation setting called Generalized Domain Adaptation (GDA) which possesses the following characteristics: 1) Partially or noisy labeled source data, 2) Weakly or unlabeled target data.
\begin{figure}
  \centering
  \includegraphics[width=0.48\textwidth]{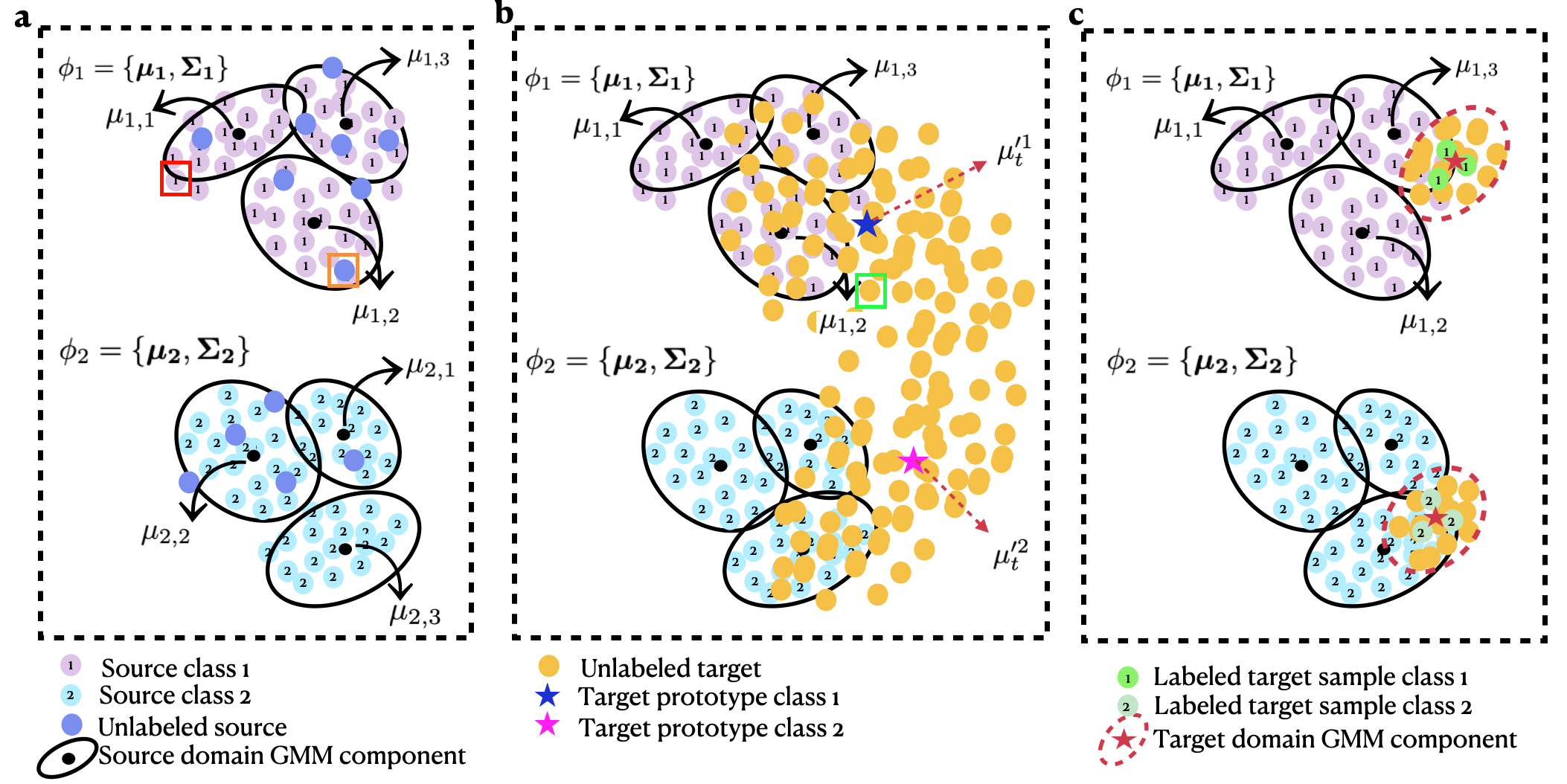} %
  \caption{GMM-based contrastive learning. a) Labeled and unlabeled source data along with the GMM model with 3 components fitted on labeled source data. b) Unlabeled target data, together with the source GMM model. c) Unlabeled target data, coupled with the adaptive GMM model fitted to labeled target data and the source GMM model.}
  \label{fig:approach}
\end{figure}
The GDA setting relaxes the problem of UDA by allowing the use of unlabeled or weakly labeled data from the source domain and weak labels from the target domain. Nevertheless, effectively leveraging these unlabeled or weakly labeled source data and weakly labeled target data is a non-trivial task. There is limited research that focuses on the incorporation of weak labels from the target domain. 
Even though \cite{akata2022urban} and \cite{paul2020domain} introduce weak labels from the target domain as supplementary sources of supervision.
In contrast, \cite{das2023weakly} utilize class prototypes generated by exploiting these weak labels to reduce the domain gap. However, prototype-based approaches, relying on the unimodality assumption within each category, may not fully account for variations in specific attributes  (e.g. shape) \cite{liang2022gmmseg}. This oversight could potentially diminish the distinguishability of the acquired features. Furthermore, despite limited studies on weekly supervised domain adaptation, the incorporation of unlabeled/weakly labeled source data remains an underexplored area. 

This work addresses domain adaptation challenges in GDA settings.
We propose a Generalized Gaussian-mixture-based (GenGMM) Domain Adaptation Model, leveraging the source and target domain distributions to enhance the quality of weak and pseudo labels and achieve alignment between the source and target domains.
The GenGMM approach is based on two key principles: 1) the domain closeness assumption \cite{shi2012information} and 2) feature similarity between labeled and unlabeled pixels \cite{wu2023sparsely}. 
Expanding on these principles, GenGMM diverges from the common approach of relying on unreliable pseudo-labels generated by the discriminative classifier for contrastive learning. Instead, it identifies positive and negative clusters for a given target sample by considering the underlying distribution of both source and target domains. 
We estimate the source domain's underlying distribution using the generative GMM.
There are three distinct advantages in estimating the source pixel feature distribution using the GMM. First, it adapts effectively to multimodal data densities. Second, by capturing category-wise Gaussian mixtures for feature representations, the components of the GMM can serve as the most suitable representative prototypes for contrastive loss to align the source and target domain. Third, the GMM model provides reliable information from labeled pixels to refine the unlabeled or weakly labeled pixels, achieving more reliable supervision on both source and target domains. 
Inspired by \cite{wu2023sparsely}, we model the similarity between the labeled and unlabeled pixels of the target domain by using the adaptive GMM. 
This enables us to utilize the soft GMM predictions to provide more probabilistic guidance for unlabeled regions, enabling more guided and effective contrastive learning. 
We summarise our main contributions as:
1- We formally introduce the GDA setting, where both the source and target domains are weakly labeled.
2- We propose the GenGMM model, which incorporates the underlying distribution of the source and target domain data using a GMM as a generative model, in conjunction with a discriminative classifier, to enhance the performance of the UDA model.
3- Extensive experiments conducted on numerous benchmark datasets have confirmed the effectiveness of the GenGMM approach within the GDA settings.
4- The experimental results indicate that the GenGMM approach achieved high performance where the source data contains real-world label noise. 

%% file: methodology.tex
\begin{algorithm}[!t]
\caption{GenGMM model}  
\label{alg:algorithm}
\textbf{Input}:$S_{l}$,  $S_{u}$, $T_{l}$, and $T_{u}$
\begin{algorithmic}[1] 
\FOR{Iter = 1: $N_{Iter}$}
\FOR{$n\in{1,...,N_{batch}^{s,l}}$(Labeled source minibatch)}
\STATE Get pixel source features $f_{s}^{l}$
\STATE Update source pixel data distribution GMM model ${\{\phi^{*}_{c}\}}$ using Sinkhorn EM(Sec. \ref{sec:source_dist})
\STATE Identify positive\&negative prototypes(Eq. \ref{eq:4}-\ref{eq:5})
\STATE Set $\alpha=1$ for pixel feature $f_{s}^{l}$
\STATE Compute loss $l_{GMMCl}$ for feature $f_{s}^{l}$ (Eq.\ref{eq:3})
\ENDFOR
\FOR{$n\in{1,...,N_{batch}^{s,u}}$(Un/weakly labeled source minibatch)}
\STATE \textbf{Get pixel source features $f_{s}^{u}$}
\STATE Determine positive\&negative prototypes (Eq.\ref{eq:6})
\STATE Compute $\alpha$ for pixel feature $f_{s}^{u}$  (Eq.\ref{eq:7})
\STATE Compute loss $l_{GMMCl}$ for feature $f_{s}^{u}$  (Eq.\ref{eq:3})
\ENDFOR
\FOR{$n\in{1,...,N_{batch}^{t}}$(Target minibatch)}
\STATE Get pixel target features $f_{t}$
\STATE Assign pseudo-label to unlabeled data (Eq. \ref{eq:8}-\ref{eq:9})
\STATE Determine positive\&negative prototypes(Eq.\ref{eq:10})
\IF{Contains weak labeled target data}
\STATE Estimate target GMM via weak labels (Eq. \ref{eq:11})
\STATE Compute $\alpha$ for pixel feature $f_{t}$  (Eq.\ref{eq:12})
\ELSIF{Contains only unlabeled target data}
\STATE Set $\alpha=1$ for pixel feature $f_{t}$ 
\ENDIF
\STATE Compute loss $l_{GMMCl}$ for feature $f_{t}$  (Eq.\ref{eq:3})
\ENDFOR
\ENDFOR
\end{algorithmic}
\end{algorithm}
In this section, we start with an overview of the foundations (Sec. \ref{sec:Preliminaries}), introducing GDA settings and essential preliminary information. We then explore the key components of the GenGMM approach, beginning with estimating the source domain distribution (Sec. \ref{sec:source_dist}) by using the GMM. Next, we explain our GMM-based contrastive learning approach (Sec. \ref{sec:gmmcl}), outlining our strategy for minimizing the domain gap between the source and target domains using contrastive learning, guided by the pixel distribution of both domains.
\subsection{Preliminaries}
\label{sec:Preliminaries}
The source and target domain distributions are $p_{s}(x,y) \in p_{\mathcal{S}}$ and $p_{t}(x,y) \in p_{\mathcal{T}}$, from which the source $\mathcal{D}_{S}$ and target  $\mathcal{D}_{T}$ are sampled i.i.d. We sample labeled source data $S_{l}=\{x_{i}^{s,l},y_{i}^{s,l} \}$ from $\mathcal{D}_{S}$, unlabeled source data $S_{u}=\{x_{i}^{s,u}\}$ from the marginal distribution of $\mathcal{D}_{S}$, labeled target data $T_{l}=\{x_{i}^{t,l},y_{i}^{t,l} \}$ from $\mathcal{D}_{T}$, unlabeled target data $T_{u}=\{x_{i}^{t,u}\}$ from the marginal distribution of $\mathcal{D}_{T}$, where $x \in R^{H \times W \times 3}$, $y \in R^{H \times W \times C}$ and $C$ is the number of classes.
\begin{equation}
\begin{aligned}    
L_{ce}^{l}= -\underset{d\in\{s,t\}}{\sum}\overset{H\times W}{\underset{i=1}{\sum}}\overset{C}{\underset{c=1}{\sum}} I_{[y^{d,l}_{i}=c]}log(\hat y^{d,l}_{i,c})
\label{eq:1-1}
\end{aligned}
\end{equation}
\begin{equation}
\begin{aligned}    
L_{ce}^{u}&= -\underset{d\in\{s,t\}}{\sum}\overset{H\times W}{\underset{i=1}{\sum}}\overset{C}{\underset{c=1}{\sum}} w_{t,i,c}I_{[\hat{y}^{d,u}_{i}=c]}log(\hat y^{d,u}_{i,c})\\
\hat{y}^{d,u}_{i}&=\underset{c}{argmax}\:\:\hat y^{d,u}_{i,c}, \:\:\:\:\:\:\:\:\: I \in \{1,2,..,H\times W\}\\
w_{t,i,c}&=\frac{\overset{H\times W}{\underset{i=1}{\sum}}1_{[\underset{c}{max}\:\:\hat y^{d,u}_{i,c}]>\beta}}{H\times W}
\label{eq:1-2}
\end{aligned}
\end{equation}
To enhance the model's performance in the target domain, GDA primarily focuses on training with all $S_{l}$, $S_{u}$, $T_{l}$, and $T_{u}$. The model itself is comprised of three key elements: an encoder (E), a multi-class segmentation head (CL), and an auxiliary projection head (F). When given an input image $x$, the auxiliary projection head processes the encoder's output to generate a feature map ($f = F(E(x))$). These features are then transformed into an $l_{2}$-normalized feature vector. Subsequently, the multi-class segmentation head operates on the encoder's output to produce a class probability map ($\hat y = CL(E(x))$). For model training, we employ the cross-entropy loss ($L_{ce}$) over both labeled and unlabeled data (i.e., Eqs. \ref{eq:1-1} - \ref{eq:1-2}), in addition to the GenGMM loss functions. Eq. \ref{eq:1-1} represents the cross-entropy loss function applied to the labeled data from both the source and target domains, utilizing the ground truth labels $y^{d,l}_{i}$, with $d$ indicating the domain, which can take values source ($s$) or target ($t$). However, the cross-entropy loss function for unlabeled data from both the source and target domains is computed using Eq. \ref{eq:1-2}. Since pseudo labels are typically noisy, as suggested in \cite{xie2023sepico}, we apply weights ($w_{t,i,c}$) to the loss values by using Eq. \ref{eq:1-2}. In addition, we implement the teacher-student architecture \cite{tarvainen2017mean}.

\subsection{Source domain distribution}
\label{sec:source_dist}

Our approach focuses on estimating the source domain distribution, denoted as $p(f_{s},c)=p(f_{s}|c)p(c)$, using labeled source data. To achieve this, we estimate two vital components: the class conditional distribution $p(f_{s}|c)$ and the class prior $p(c)$. We establish uniform class source priors using rare class sampling (RCS), a technique introduced in DAFormer \cite{hoyer2022daformer}. Additionally, as shown in Figure \ref{fig:approach}-a, we employ a GMM model that consists of a weighted mixture of M multivariate Gaussians to estimate the class-conditional distribution $p(f_{s}|c;\phi_{c})$ for each category $c$ ($Eq. \ref{eq:2}$). The GMM classifier, parameterized as ${\phi_{c}=\{\boldsymbol{\pi_{c}}, \boldsymbol{\mu_{c}}, \boldsymbol{\Sigma_{c}}\}}_{c=1}^{C}$, is  optimized online using a momentum-based variant of the Sinkhorn Expectation-Maximization algorithm, as proposed by \cite{liang2022gmmseg}. The GMM model is optimized exclusively using source pixel embeddings with pseudo labels that match their ground truth labels. Consequently, the model remains unaffected by real-world label noise in noisy source domains.
\begin{equation}
\begin{aligned}    p(\boldsymbol{f_{s}}|c;\boldsymbol{\phi_{c}})=\Sigma_{m=1}^{M}\pi_{c,m}\mathcal{N}(\boldsymbol{f_{s}};\boldsymbol{\mu_{c,m}},\boldsymbol{\Sigma_{c,m}}) 
\label{eq:2}
\end{aligned}
\end{equation}
where, $\pi_{c,m}$ represents the prior probability for each class, with the constraint that $\Sigma_{m=1}^{M}\pi_{cm}=1$. $\boldsymbol{\Sigma_{c}}$ and $\boldsymbol{\mu_{c}}$ signifies the covariance matrix and mean vector.

\subsection{GMM-based contrastive learning}
\label{sec:gmmcl}
We perform contrastive learning between the pixel embeddings ($f_{d}, \forall d \in \{s,t\}$) and the components of the source GMM model, which act as the most representative prototypes, using Eqs. \ref{eq:3}. Nevertheless, the selection of negative ($q^{-}$) and positive ($q^{+}$) prototypes is challenging for unlabeled or weakly labeled pixels, as pseudo-labels from the discriminative classifier are noisy.  \cite{vayyat2022cluda}.
\begin{equation}
\begin{aligned}  
l_{GMMCl} =-\alpha log \frac{e^{f_{d}q^{+}_{c}/\tau}}{e^{f_{d}q^{+}_{c}/\tau}+\underset{c' \neq c}{\Sigma_{c=1}^{C}}e^{f_{d}q^{-}_{c'}}/\tau}
\label{eq:3}
\end{aligned}
\end{equation}
To address this challenge, we leverage two foundational assumptions: 1) the domain closeness assumption \cite{shi2012information} and 2) the feature similarity between labeled and unlabeled pixels \cite{wu2023sparsely}. According to these two assumptions, we propose to select the positive/negative prototypes based on the similarity between unlabeled or weakly labeled pixel embeddings and prototypes, which is measured using the fitted GMM model on the labeled source pixel embeddings. This, in turn, allows the method to identify positive/negative samples with greater precision. To further mitigate noise, we introduce weighting ($\alpha$) that considers each feature embedding's proximity to the positive prototypes in contrastive training. The details of choosing the positive/negative prototypes for source and target pixel embeddings, along with the value of $\alpha$, are described in the following sections. The summary of the GenGMM model is shown in Algorithm \ref{alg:algorithm}. 
 
\subsubsection{Labeled source pixel embeddings}

Based on the values of $p_{s}(m|f_{s},c;\phi^{}_{c})$ for the given source sample and its corresponding ground truth label $c$, we determine both positive and negative prototypes for labeled pixels. Here, $p_{s}(m|f_{s},c;\phi^{}_{c})$ represents the posterior probability, signifying the likelihood of data $f_{s}$ being assigned to component $m$ within class $c$. We compute this probability using Bayes' rule while assuming uniform class priors, as detailed in Eq. \ref{eq:4}. We choose positive prototypes by taking the prototype associated with the mean of the nearest component that shares the same label, as indicated in Eq. \ref{eq:5}. In parallel, to find the hardest negative prototypes across various categories with distinct labels, we identify the closest component per category, following the procedure described in Eq. \ref{eq:5}. Noted, the number of hardest negative prototypes for each pixel embedding is C-1. For example, as depicted in Figure \ref{fig:approach}-a, we select $\mu_{1,1}$ and $\mu_{2,2}$ as the positive and negative prototypes for the labeled source sample with a ground truth label of 1 in the red square. Contrastive training is conducted using Eq. \ref{eq:3}, with $\alpha$ set to 1.
\begin{equation}
\begin{aligned}    
 p_{s}(m|f_{s},c;\phi^{*}_{c})=\frac{\pi_{c,m}\mathcal{N}(f_{s}|\mu_{c,m},\Sigma_{c,m})}{\Sigma_{m'=1}^{M}\pi_{c,m'}\mathcal{N}(f_{s}|\mu_{c,m'},\Sigma_{c,m'})}
\label{eq:4}
\end{aligned}
\end{equation}
\begin{equation}
\begin{aligned}    
q^{+}&=\{\mu_{c,m^{+}}|\:\:m^{+}=\underset{m}{\arg\max}\:\:p_{s}(m|f_{s},c;\phi^{*}_{c}), c=y^{s,l}\}\\
q^{-}_{c}&=\{\mu_{c,m^{-}}|\:\:m^{-}=\underset{m}{\arg\max}\:\:p_{s}(m|f_{s},c;\phi^{*}_{c}),c\}, \forall c \neq y^{s,l}
\label{eq:5}
\end{aligned}
\end{equation}
\subsubsection{Unlabeled source pixel embeddings}

\label{sec:source_unlabeled}
Regarding the unlabeled source pixels ($S_{u}=\{x_{i}^{s,u}\}$), they pose a challenge as they lack any available ground truth labels for determining positive and negative prototypes. A common method is to assign pseudo labels to $X^{s,u}$ using the output of a discriminative classifier. However, these pseudo-labels are often noisy. 
We propose an effective approach for guiding unlabeled pixels. Given that pixels with identical semantic labels tend to cluster together in the feature space \cite{wu2023sparsely}, we suggest leveraging the GMM model fitted to labeled source data in Sec. \ref{sec:source_dist} to model the similarity between labeled and unlabeled pixels. By utilizing the GMM model and measuring the similarity of a given pixel's embedding with the components of the GMM model, we select positive and hardest negative components as depicted in Eq. \ref{eq:6}.
For instance, in Figure \ref{fig:approach}-a, we choose $\mu_{1,2}$ as the positive prototype and $\mu_{2,1}$ as the hardest negative prototype for the unlabeled source sample in the orange square. 
To enhance the training process, we introduce a weighting mechanism for the contrastive training loss associated with each pixel embedding. This weight, denoted as $\alpha$ and calculated using the closest GMM component with the same labels as described in Eq. \ref{eq:7}. 
\begin{equation}
\begin{aligned}  
&q^{+}=\{\mu_{c^{+},m^{+}}|\:\:c^{+},m^{+}=\underset{c,m}{\arg\max}\:\:p_{s}(c,m|f_{s},\phi_{c})\}\\
&q^{-}_{c}=\{\mu_{c,m^{-}}|\:\:m^{-}=\underset{m}{\arg\max}\:\:p_{s}(m|f_{s},c;\phi^{*}_{c}),c\},  \forall c \neq c^{+}
\label{eq:6}
\end{aligned}
\end{equation}
\begin{equation}
\alpha=e^{-\frac{d^{2}}{2\sigma_{c^{+},m^{+}}}}
\label{eq:7}
\end{equation}
where, $d$ represents the difference between $f_{s}$ and $\mu_{c^{+},m^{+}}$. Weighting the contrastive training loss is essential to mitigate the impact of noise, as it can significantly affect contrastive training \cite{vayyat2022cluda}. Notably, by omitting the term $\frac{1}{\sqrt{2\pi \sigma^2}}$, we restrict $\alpha$ to the range of 0 to 1. The value of $\alpha$ indicates the proximity of a given pixel to its associated GMM component, with higher values indicating greater similarity. 

\subsubsection{Noisy source-labeled pixel embeddings}
In the presence of noisy source-labeled data, we apply weighted contrastive training to alleviate the impact of noise, following the same framework as described in Sec. \ref{sec:source_unlabeled}. As detailed in Sec. \ref{sec:source_unlabeled} the weights ($\alpha$) are computed based on the closest GMM component with the same label as the given pixel embedding. Despite the noise in the source domain data, the source GMM model remains reliable. This reliability is due to its construction process, detailed in Sec. \ref{sec:source_dist}, which involves using pixel embeddings with pseudo-labels that match their ground truth labels, mitigating label noise.

\subsubsection{Unlabeled target pixel embeddings}
To determine the positive and hardest negative prototypes for the unlabeled target pixels, we propose to assign pseudo-labels to them using the posterior probability $p_{t}(c|f_{t};\phi^{*}_{c})$ and their similarities to the target prototypes, as demonstrated in Eq. \ref{eq:8}. 
The key concern is how to derive the target domain prototypes and determine the posterior probability for a given target embedding. Target domain prototypes per category are established through an exponential moving average, utilizing reliable pixel embeddings sourced from the target memory bank. This iterative process involves updating the target bank with reliable pixel embeddings from target mini-batches. It begins by computing class-specific average pixel embeddings based on pseudo labels and assessing their cosine similarity with class means. Subsequently, the M pixel embeddings with the highest cosine similarity scores, signifying their trustworthiness, are chosen and incorporated into the target bank. 
The posterior probability is computed using Proposition 1. The rationale behind utilizing Eq. \ref{eq:8} for reliable pseudo-label generation is rooted in the assumption that features from both domains cluster together in a shared space. As depicted in Figure \ref{fig:approach}-b, the unlabeled target sample within the green square is close to target prototype class 1 and the source GMM components of class 1.

\begin{equation}
\hat y_{t}^{c}=\underset{c}{\arg\max}\:\:\:p_{t}(c|f_{t};\phi^{*}_{c})\times \frac{e^{cosine(\mu^{t}_{c},f_{t})}}{\Sigma_{c'}e^{cosine(\mu^{t}_{c'},f_{t})}}
\label{eq:8}
\end{equation}
\begin{equation}
\begin{aligned}  
&p_{t}(c|f_{t};\phi^{*}_{c})=\underset{m'}{\sum}{p_{s}(c,m'|f_{t};\phi^{*}_{c})} \times \frac{\delta_{target}^{c}}{\delta_{source}^{c}}
\label{eq:9}
\end{aligned}
\end{equation}
\begin{equation}
\begin{aligned}    
q^{+}=\{\mu_{c,m^{+}}|\:\:m^{+}=\underset{m}{\arg\max}\:\:p_{t}(m|f_{t},c;\phi^{*}_{c})\\
, c=\hat y^{t,u}\}\\
q^{-}_{c}=\{\mu_{c,m^{-}}|\:\:m^{-}=\underset{m}{\arg\max}\:\:p_{t}(m|f_{t},c;\phi^{*}_{c})\\
,c\} \forall c \neq \hat y^{t,u}&
\label{eq:10}
\end{aligned}
\end{equation}
\textbf{Proposition 1:} Given $p_{s}(c|f_{t};\phi^{*}_{c})=\underset{m'}{\sum}{p_{s}(c,m'|f_{t};\phi^{*}_{c})}$, the posterior probability for the given target sample $f_{t}$ is:

\begin{equation}
\begin{aligned}  
p_{t}(c|f_{t};\phi^{*}_{c})&=p_{s}(c|f_{t};\phi^{*}_{c})\times \frac{\delta_{target}^{c}}{\delta_{source}^{c}}
\label{eq:12}
\end{aligned}
\end{equation}
Noted, adjusting the posterior using the ratio $\frac{\delta_{target}^{c}}{\delta_{source}^{c}}$ addresses the issue of label shift. $\delta_{target}^{c}$ and $\delta_{source}^{c}$ are the prior distribution of the target and source domain and are updated using the exponential moving average during training.

\textit{proof:} Based on the Bayes rule, We have the below relationship for the source and target posterior probabilities: 
\begin{equation}
\begin{aligned}  
p_{t}(c|f_{t};\phi^{*}_{c})\:&\alpha \:p_{t}(f(x)|c;\phi^{*}_{c})p_{t}(c)\\
p_{s}(c|f_{t};\phi^{*}_{c})\:&\alpha \:p_{s}(f(x)|c;\phi^{*}_{c})p_{s}(c)
\label{eq:13}
\end{aligned}
\end{equation}

If we assume the conditional data distribution is well aligned, i.e. $p_{t}(f(x)|c)=p_{s}(f(x)|c)$. We can extract the below relationship between the posterior probabilities:  
\begin{equation}
\begin{aligned}  
p_{t}(c|f_{t};\phi^{*}_{c})\:=
p_{s}(c|f_{t};\phi^{*}_{c})\times \frac{p_{t}(c)}{p_{s}(c)}\:
=p_{s}(c|f_{t};\phi^{*}_{c})\times \frac{\delta_{target}^{c}}{\delta_{source}^{c}}
\label{eq:14}
\end{aligned}
\end{equation}
Finally, the contrastive learning loss value is computed using Eq. \ref{eq:3} with $\alpha =1$, given $q^{+}$ and $q^{-}_{c}$ via Eq. \ref{eq:10}.


\subsubsection{Weak labeled target pixel embeddings}
\label{sec:weaktarget}
We enhance model performance through refined contrastive learning using weak labels. This approach relies on the assumption that the unlabeled target data is close to both the labeled source data and the labeled target data with the same semantic label \cite{wu2023sparsely}. First, drawing on our knowledge of the source data distribution (i.e. Eq. \ref{eq:2}), we determine the pseudo-labels, positive and hard negative prototypes using Eqs. \ref{eq:8}-\ref{eq:10}. 
Then, in cases of point or coarse annotations for target-domain images, we mitigate the influence of noisy pseudo-labels during training based on the assumption of the closeness of unlabeled target data to the labeled target data. Inspired from \cite{wu2023sparsely}, we model this similarity by fitting a GMM with K components to each target image, where K represents the number of classes, as shown in Figure \ref{fig:approach}-c. Mean and covariance are computed for each GMM component, as follows: 
\begin{equation}
\begin{aligned}  
\mu_{k}&=\frac{1}{\Sigma_{I_{y_{n}^{t,l}=k}}}\Sigma_{n}I_{{y_{n}^{t,l}=k}}f_{t,n}\:\:\:\:\:\:\:\:\:\:\ \forall k=y^{t,l}\\
\sigma_{k}&=\sqrt{\frac{1}{N}\Sigma_{n=1}^{N}\hat y^{t}_{n}(f_{t,n}-\mu_{k})^{2}}
\label{eq:11}
\end{aligned}
\end{equation}
\begin{equation}
\alpha=e^{-\frac{d^{2}}{2\sigma_{k}}}\:\:\:\:\:\:\:\forall k=m^{+}
\label{eq:12}
\end{equation}
We then compute $\alpha$ values using Equation \ref{eq:12} as weighting factors to adjust each pixel's contribution during contrastive learning through Eq. \ref{eq:3}. 
Lower $\alpha$ values indicate weaker proximity to pixels of the same class, allowing us to reduce their impact on the loss function. Also, in the presence of weak target data, the $\alpha$ is used to weight the self-training instead of $w$, as it is more informative (See Sec. 2.1).
\begin{table*}[h]
  \centering
  \begin{tabular}{@{}p{2.2cm}@{}*{20}{|@{\hspace{.7pt}}c@{}}}
    \toprule
    \multicolumn{21}{c}{\textbf{Cityscapes}$\rightarrow$\textbf{Dark Zurich}}\\
    \hline
    Model&	Road&	S.Wa&	Bld.&	Wall&Fence&	Pole&	T.Lig&T.Sig&	Veget.&	Ter.&Sky&	Pers.&	Rider&	Car&	Truck&	Bus&Train&	M.Bike&	Bike&	mIoU\\
    \hline
    DAFormer\cite{hoyer2022daformer}&84.8&47.2&66.5&35.0&13.3&\textbf{45.4}&14.4&\textbf{34.8}&48.1&\textbf{27.4}&62.2&48.9&44.5&63.3&52.6&\textbf{0.8}&83.7&\textbf{40}&35&44.6\\
    \hline
    SePiCo\cite{xie2023sepico}&\textbf{88.1}&\textbf{54.7}&67.8&31.9&18.3&41.0&24.6&32.4&59.7&21.5&78.3&34.2&\textbf{45.1}&68.3&33.5&0&26.8&14.3&16.9&39.9\\
    \hline
    GenGMM&86.9&45.1&\textbf{73.5}&\textbf{41.0}&\textbf{19.7}0&21.8&\textbf{42.0}&29.9&\textbf{66.3}&21.9&\textbf{79.3}&\textbf{54.7}&36.6&\textbf{76.4}&\textbf{75.8}&0.4&\textbf{85.2}&38.6&\textbf{38.6}&\textbf{49.0}\\
    \bottomrule
  \end{tabular}
  \caption{Comparison with state-of-the-art methods for noisy labeled source data}
  \label{table:noisy}
\end{table*}
\begin{table}
  \centering
  \begin{tabular}{@{}p{2.2cm}@{}*{6}{|@{}c@{}}}
    \toprule
    Data&\multicolumn{3}{c|}{\textbf{GTA5}$\rightarrow$\textbf{City.}}&\multicolumn{3}{c}{\textbf{Synthia}$\rightarrow$\textbf{City.}}\\
    \hline
    Model&	50\% &	70\% &	100\% &	50\% &	70\% &	100\% \\
    \hline
    DAFormer\cite{hoyer2022daformer}&65.5&65.4&68.3&58.2&59.1&60.9\\
    \hline
    SePiCo\cite{xie2023sepico}&63.8&65.0& 69.7&59.7&60.5&62.2\\
    \hline
    GenGMM&\textbf{67.8}&\textbf{68.3}&\textbf{70.4}&\textbf{61.4}&\textbf{62.0}&\textbf{63.3}\\
    \bottomrule
  \end{tabular}
  \caption{Comparison for partially labeled source data}
  \label{table:50-50}
\end{table}

%% file: experiments.tex
\begin{figure}
  \centering  \includegraphics[width=0.9\linewidth]{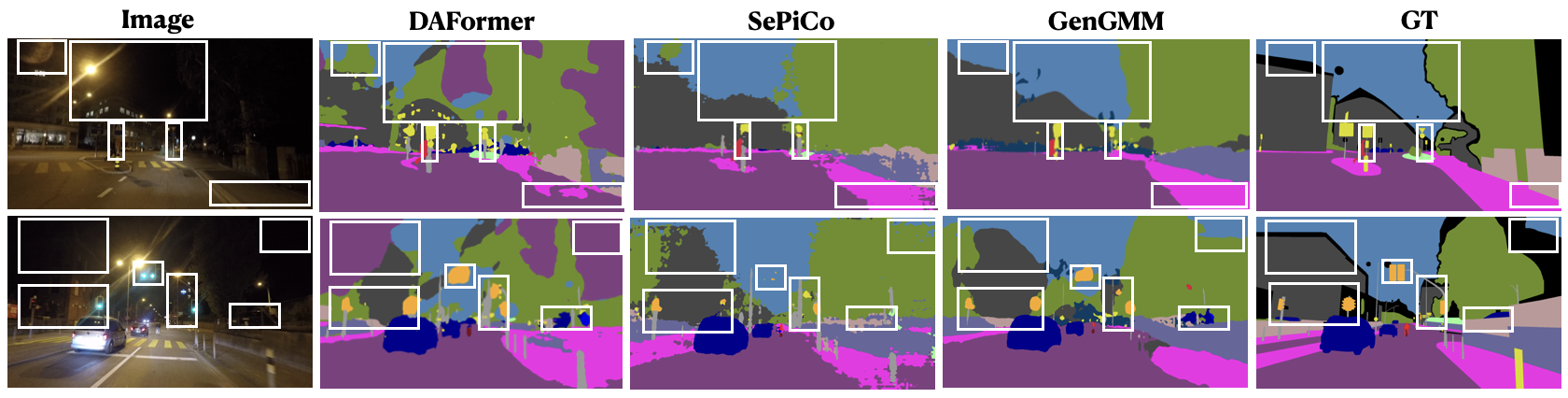}
    \caption{Qualitative results on Cityscapes→Dark Zurich}
    \label{fig:zurich}
\end{figure}
\begin{table}
  \centering
   \begin{tabular}{p{2.2cm}|*{2}{@{\hspace{1pt}}c@{\hspace{1pt}}}|*{1}{@{}c@{}}*{2}{@{\hspace{4pt}}c@{\hspace{2pt}}}}
    \toprule
    Data&\multicolumn{2}{c|}{\textbf{GTA5}$\rightarrow$\textbf{City.}} & \multicolumn{3}{c}{\textbf{Synthia}$\rightarrow$\textbf{City.}}\\
    \hline
    Model&	mIoU&	gap& mIoU&mIoU\textsuperscript{*}&gap\\
    \hline
    WeakSegDA\cite{paul2020domain}&56.4&-&57.2&63.7&-\\
    \hline
    WDASS\cite{das2023weakly}&64.7&+8.3&62.8&68.7&+5.0\\
    \hline
    GenGMM&\textbf{71.4}&+6.7&\textbf{65.1}&\textbf{72.4}&+3.4\\
    \bottomrule
  \end{tabular}
  \caption{The comparison with SOTA methods for Point labels}
  \label{table:point}
\end{table}
\begin{table}
  \centering
   \begin{tabular}{p{2.7cm}|*{2}{@{\hspace{2pt}}c@{\hspace{2pt}}}|*{1}{@{}c@{}}*{2}{@{\hspace{1pt}}c@{\hspace{1pt}}}}
    \toprule
    Data&\multicolumn{2}{c|}{\textbf{GTA5}$\rightarrow$\textbf{City.}} & \multicolumn{3}{c}{\textbf{Synthia}$\rightarrow$\textbf{City.}}\\
    \hline
    Model&	mIoU&	gap& mIoU & mIoU\textsuperscript{*}&gap\\
    \hline
    Coarse-to-fine\cite{das2023urban}&66.7&-&61.6&67.2&-\\
    \hline
    WDASS\cite{das2023weakly}&69.1&+2.4&66.0&71.0&+3.8\\
    \hline
    GenGMM&\textbf{72.3}&+3.2&\textbf{71.6}&\textbf{76.2}&+5.2\\
    \bottomrule
  \end{tabular}
  \caption{The comparison with SOTA methods (Coarse labels)}
  \label{table:coarse}
\end{table}
\begin{table}
  \centering
  \begin{tabular}{@{}lc|cc@{}}
    \toprule
    Model&\multicolumn{1}{c}{\textbf{GTA5}$\rightarrow$\textbf{City.}} & \multicolumn{1}{c}{\textbf{Synthia}$\rightarrow$\textbf{City.}}\\
    \hline
    \multicolumn{3}{c}{\textbf{Point annotations}}\\
    \hline
    Baseline (DAFormer)&68.9&61.9\\
    \hline
    GenGMM&\textbf{71.4}&\textbf{65.1}\\
    \hline
    \multicolumn{3}{c}{\textbf{Coarse annotations}}\\
    \hline
    Baseline (DAFormer)&69.2&69.6\\
    \hline
    GenGMM&\textbf{72.3}&\textbf{71.6}\\
    \bottomrule
  \end{tabular}
  \caption{The comparison with the Baseline model}
  \label{table:resultbaseline}
\end{table}
\begin{table}
  \centering
  \begin{tabular}{@{}c@{}|@{}c@{}|@{}c@{}|*{2}{@{}c@{}}}
    \toprule
    Lb&UL&GMM-Cl&\multicolumn{1}{c}{\textbf{GTA5}$\rightarrow$\textbf{City.}} & \multicolumn{1}{c}{\textbf{Synthia}$\rightarrow$\textbf{City.}}\\
    \hline
    \checkmark&&&64.7&60.0\\
    \hline
    \checkmark&\checkmark&&65.1&60.5\\
    \hline
    \checkmark&\checkmark&\checkmark&\textbf{67.8} &\textbf{61.4}\\
    \bottomrule
  \end{tabular}
  \caption{Comparison wrt to components}
  \label{table:ablation1}
\end{table}
\begin{table}
  \centering
   \begin{tabular}{p{2.2cm}|*{2}{@{\hspace{1pt}}c@{\hspace{1pt}}}|*{1}{@{}c@{}}*{1}{@{\hspace{4pt}}c@{\hspace{2pt}}}}
    \toprule
    Data&\multicolumn{2}{c|}{\textbf{GTA5}$\rightarrow$\textbf{City.}} & \multicolumn{2}{c}{\textbf{Synthia}$\rightarrow$\textbf{City.}}\\
    \hline
    Model&	$w$&	$\alpha$& $w$&	$\alpha$\\
    \hline
    GenGMM&70.5&\textbf{71.4}&63.6&\textbf{65.1}\\
    \bottomrule
  \end{tabular}
  \caption{The effect of $\alpha$ (point annotations)}
  \label{table:weight}
\end{table}
\begin{table}
  \centering
  \begin{tabular}{*{1}{@{}c@{}}*{8}{|@{}c@{}}}
    \toprule
    Model&\multicolumn{4}{c}{\textbf{GTA5}$\rightarrow$\textbf{City.}} & \multicolumn{4}{c}{\textbf{Synthia}$\rightarrow$\textbf{City.}}\\
    \hline
    M&1&3&5&7&1&3&5&7\\
    \hline
    mIoU&69.2&69.7&\textbf{70.4}&69.6&61.7&\textbf{63.3}&62.9&61.2\\
    \bottomrule
  \end{tabular}
  \caption{Number of components}
  \label{table:components}
\end{table}

\subsection{Datasets and metric}
We conducted our evaluation on four well-established domain adaptation benchmark datasets: GTA5\cite{richter2016playing}, Cityscapes\cite{cordts2016cityscapes}, SYNTHIA\cite{ros2016synthia}, and Dark Zurich\cite{sakaridis2019guided}. 
For Cityscapes, we used 2975 training images and reported results on a validation dataset of 500 images. In the case of GTA5, Synthia, and Dark Zurich, we had 24,966, 9,400, and 2,416 training images, respectively.
We evaluated our approaches based on the mean Intersection over Union (mIoU) score \cite{xie2023sepico}.

\subsection{Network architecture and training}

Our network architecture is based on the DAFormer framework \cite{hoyer2022daformer}, incorporating the Class-balanced Cropping (CBC) approach and regularization methods from \cite{xie2023sepico}. 
We initialize the backbone model with pre-trained ImageNet weights \cite{deng2009imagenet} and add two $1 \times 1$ convolutional layers with ReLU activation \cite{xie2023sepico} to obtain a 64-dimensional feature vector, which is further $l_2$-normalized (D=64). 
Our model was constructed utilizing PyTorch version 1.8.1, and trained on a single NVIDIA Tesla V100-32G GPU. The optimization strategy is AdamW \cite{loshchilov2017decoupled} with betas (0.9, 0.999) and a weight decay of 0.01. Learning rates are $6 \times 10^-5$ for the encoder and $6 \times 10^-4$ for the decoder. All Exponential Moving Averages update weights are set to 0.9, except for the teacher network ($\beta=0.999$).
The number of epochs and batch size are 60,000 and 2. Notably, we impose a diagonal structure constraint on the covariance matrices $\Sigma \in \mathbb{R}^{D \times D}$ used in our GMM model. 
To optimize the performance of our GMM, we implement a generative optimization process, proposed in the GMMSeg framework \cite{liang2022gmmseg}. In each iteration the momentum Sinkhorn Expectation-Maximization process is executed. The number of components is 5 for GTA5 and Dark Zurich and 3 for Synthia. The external memory size is 32k per category. This memory is updated on a first-in, first-out basis, with 100 pixels per class. 

\subsection{Comparison with existing UDA methods}
We perform three separate comparisons:  1) noisy labeled source data, 2) partially labeled source data, and 3) weakly labeled target data. In the first scenario, where the source data exhibits noise, we assess the performance of the GenGMM model using the Cityscapes $\rightarrow$ Dark Zurich. To replicate the noisy source conditions, we employed coarse annotations from the Cityscapes dataset as training labels, representing real-world label noise. Our evaluation, detailed in Table \ref{table:noisy}, compares the GenGMM model against the DAFormer \cite{hoyer2022daformer} and SePiCo \cite{xie2023sepico} models. Notably, our GenGMM method outperforms both the DAFormer and SePiCo models by a margin of 4.4\% and 9.1\%.
Table \ref{table:50-50} presents the results for the second comparison, assuming scenarios where 50\%, 70\%, and 100\% of the source data are labeled. 
Both the DAFormer and SePiCo models were trained by incorporating self-training on the unlabeled source data. As observed, our GenGMM method outperforms both DAFormer \cite{hoyer2022daformer} and SePiCo \cite{xie2023sepico} models for both GTA5 $\rightarrow$ Cityscapes and Synthia $\rightarrow$ Cityscapes. 
In our latest comparison, we assessed the GenGMM model's performance with weakly labeled data, both point and coarse labels. Noted, we generated point labels for each class within images at their original sizes by randomly selecting a small group of pixels. This selection was done within a randomly positioned circle with a radius of 4 per category. We opted for circles instead of points because different methods may use various image sizes, and point annotations can be lost when resizing. We conducted comparisons with previous methods, namely, WeakSegDA \cite{paul2020domain} and WDASS \cite{das2023weakly} for point labels, and Coarse-to-fine \cite{das2023urban} and WDASS \cite{das2023weakly} for coarse labels. The results, as shown in Tables \ref{table:point}-\ref{table:coarse}, reveal that GenGMM outperforms the baselines. 
Table \ref{table:resultbaseline} highlights the GenGMM model's superior performance across both DA benchmarks. The qualitative comparison is shown in Fig. \ref{fig:zurich}. 

\subsection{Ablation analysis}
Table \ref{table:ablation1} highlights the positive impact of training on unlabeled data and the effectiveness of GMM-Cl on model performance. 
The impact of $\alpha$ weights in self-training, in the context of target point annotations, is presented in Table \ref{table:weight}. 
Table \ref{table:components} shows optimal values for Gaussian components (M).


%% file: conclusion.tex
We present GenGMM, a model designed for scenarios with partial or weak labeling in both source and target domains, referred to as GDA. 
GenGMM addresses GDA by utilizing weak or unlabeled data from both domains to bridge the adaptation gap. It employs GMM to capture similarities between labeled and unlabeled data in both domains, enhancing performance. Experiments show the effectiveness of GenGMM.

%% file: main.bbl
\begin{thebibliography}{10}

\bibitem{cordts2016cityscapes}
Marius Cordts, Mohamed Omran, Sebastian Ramos, Timo Rehfeld, Markus Enzweiler, Rodrigo Benenson, Uwe Franke, Stefan Roth, and Bernt Schiele,
\newblock ``The cityscapes dataset for semantic urban scene understanding,''
\newblock in {\em Proceedings of the IEEE conference on computer vision and pattern recognition}, 2016, pp. 3213--3223.

\bibitem{hoyer2022daformer}
Lukas Hoyer, Dengxin Dai, and Luc Van~Gool,
\newblock ``Daformer: Improving network architectures and training strategies for domain-adaptive semantic segmentation,''
\newblock in {\em Proceedings of the IEEE/CVF Conference on Computer Vision and Pattern Recognition}, 2022, pp. 9924--9935.

\bibitem{xie2023sepico}
Binhui Xie, Shuang Li, Mingjia Li, Chi~Harold Liu, Gao Huang, and Guoren Wang,
\newblock ``Sepico: Semantic-guided pixel contrast for domain adaptive semantic segmentation,''
\newblock {\em IEEE Transactions on Pattern Analysis and Machine Intelligence}, 2023.

\bibitem{akata2022urban}
Zeynep Akata, Bernt Schiele, Yang He, Yongqin Xian, and Anurag Das,
\newblock ``Urban scene semantic segmentation with low-cost coarse annotation,''
\newblock 2022.

\bibitem{paul2020domain}
Sujoy Paul, Yi-Hsuan Tsai, Samuel Schulter, Amit~K Roy-Chowdhury, and Manmohan Chandraker,
\newblock ``Domain adaptive semantic segmentation using weak labels,''
\newblock in {\em Computer Vision--ECCV 2020: 16th European Conference, Glasgow, UK, August 23--28, 2020, Proceedings, Part IX 16}. Springer, 2020, pp. 571--587.

\bibitem{das2023weakly}
Anurag Das, Yongqin Xian, Dengxin Dai, and Bernt Schiele,
\newblock ``Weakly-supervised domain adaptive semantic segmentation with prototypical contrastive learning,''
\newblock in {\em Proceedings of the IEEE/CVF Conference on Computer Vision and Pattern Recognition}, 2023, pp. 15434--15443.

\bibitem{liang2022gmmseg}
Chen Liang, Wenguan Wang, Jiaxu Miao, and Yi~Yang,
\newblock ``Gmmseg: Gaussian mixture based generative semantic segmentation models,''
\newblock {\em Advances in Neural Information Processing Systems}, vol. 35, pp. 31360--31375, 2022.

\bibitem{shi2012information}
Yuan Shi and Fei Sha,
\newblock ``Information-theoretical learning of discriminative clusters for unsupervised domain adaptation,''
\newblock {\em arXiv preprint arXiv:1206.6438}, 2012.

\bibitem{wu2023sparsely}
Linshan Wu, Zhun Zhong, Leyuan Fang, Xingxin He, Qiang Liu, Jiayi Ma, and Hao Chen,
\newblock ``Sparsely annotated semantic segmentation with adaptive gaussian mixtures,''
\newblock in {\em Proceedings of the IEEE/CVF Conference on Computer Vision and Pattern Recognition}, 2023, pp. 15454--15464.

\bibitem{tarvainen2017mean}
Antti Tarvainen and Harri Valpola,
\newblock ``Mean teachers are better role models: Weight-averaged consistency targets improve semi-supervised deep learning results,''
\newblock {\em Advances in neural information processing systems}, vol. 30, 2017.

\bibitem{vayyat2022cluda}
Midhun Vayyat, Jaswin Kasi, Anuraag Bhattacharya, Shuaib Ahmed, and Rahul Tallamraju,
\newblock ``Cluda: Contrastive learning in unsupervised domain adaptation for semantic segmentation,''
\newblock {\em arXiv preprint arXiv:2208.14227}, 2022.

\bibitem{das2023urban}
Anurag Das, Yongqin Xian, Yang He, Zeynep Akata, and Bernt Schiele,
\newblock ``Urban scene semantic segmentation with low-cost coarse annotation,''
\newblock in {\em Proceedings of the IEEE/CVF Winter Conference on Applications of Computer Vision}, 2023, pp. 5978--5987.

\bibitem{richter2016playing}
Stephan~R Richter, Vibhav Vineet, Stefan Roth, and Vladlen Koltun,
\newblock ``Playing for data: Ground truth from computer games,''
\newblock in {\em Computer Vision--ECCV 2016: 14th European Conference, Amsterdam, The Netherlands, October 11-14, 2016, Proceedings, Part II 14}. Springer, 2016, pp. 102--118.

\bibitem{ros2016synthia}
German Ros, Laura Sellart, Joanna Materzynska, David Vazquez, and Antonio~M Lopez,
\newblock ``The synthia dataset: A large collection of synthetic images for semantic segmentation of urban scenes,''
\newblock in {\em Proceedings of the IEEE conference on computer vision and pattern recognition}, 2016, pp. 3234--3243.

\bibitem{sakaridis2019guided}
Christos Sakaridis, Dengxin Dai, and Luc~Van Gool,
\newblock ``Guided curriculum model adaptation and uncertainty-aware evaluation for semantic nighttime image segmentation,''
\newblock in {\em Proceedings of the IEEE/CVF International Conference on Computer Vision}, 2019, pp. 7374--7383.

\bibitem{deng2009imagenet}
Jia Deng, Wei Dong, Richard Socher, Li-Jia Li, Kai Li, and Li~Fei-Fei,
\newblock ``Imagenet: A large-scale hierarchical image database,''
\newblock in {\em 2009 IEEE conference on computer vision and pattern recognition}. Ieee, 2009, pp. 248--255.

\bibitem{loshchilov2017decoupled}
Ilya Loshchilov and Frank Hutter,
\newblock ``Decoupled weight decay regularization,''
\newblock {\em arXiv preprint arXiv:1711.05101}, 2017.

\end{thebibliography}
